\documentclass[10pt, conference, compsocconf]{IEEEtran}

\usepackage[utf8]{inputenc} 
\usepackage[T1]{fontenc}    
\usepackage[bookmarks=true]{hyperref}       
\usepackage{url}            
\usepackage{booktabs}       
\usepackage{amsfonts}       
\usepackage{nicefrac}       
\usepackage{microtype}      
\usepackage{graphicx}
\usepackage[numbers]{natbib}
\usepackage{threeparttable}

\usepackage{fancyhdr}
\fancyheadoffset{0pt}
\usepackage{fixltx2e}
\usepackage{stfloats}

\title{Deep CNN-based Speech Balloon Detection and Segmentation for Comic Books}

\author{\IEEEauthorblockN{David~Dubray\IEEEauthorrefmark{1}and Jochen~Laubrock\IEEEauthorrefmark{2}}
\IEEEauthorblockA{Department of Psychology\\
University of Potsdam \\
Potsdam, Germany\\
Email: \IEEEauthorrefmark{1}ddubray@uni-potsdam.de, \IEEEauthorrefmark{2}laubrock@uni-potsdam.de}}

\begin{document}
\maketitle

\begin{abstract}
We develop a method for the automated detection and segmentation of speech balloons in comic books, including their carrier and tails. Our method is based on a deep convolutional neural network that was trained on annotated pages of the Graphic Narrative Corpus. More precisely, we are using a fully convolutional network approach inspired by the U-Net architecture, combined with a VGG-16 based encoder. The trained model delivers state-of-the-art performance with an F1-score of over 0.94. Qualitative results suggest that wiggly tails, curved corners, and even illusory contours do not pose a major problem. Furthermore, the model has learned to distinguish speech balloons from captions. We compare our model to earlier results and discuss some possible applications.
\end{abstract}


\begin{IEEEkeywords}
Convolutional neural networks; Deconvolution; Multi-layer neural network; Machine learning; Pattern recognition; Image segmentation; Document Analysis; Comics; Speech Balloons
\end{IEEEkeywords}

%
\IEEEpeerreviewmaketitle

\section{Introduction} 
\label{sec:introduction}
Comic books are an under-analyzed cultural resource that combine visual elements with textual components. Despite the common assumption that the target audience of comic books are children, their topics and story line are often elaborate, addressing a mature audience. The label ``graphic novel'' has been created to better market comic books targeting adult readers.
Comics are visual narratives with a sequential structure, and their comprehension requires a specific visual literacy from the reader. The structure of visual narratives is complex, and can be divided into modality, grammar, and meaning, each of which can be described at the level of the individual unit as well as of a sequence~\cite{Cohn:2019cn}. Most of the work from the document analysis perspective to date has focused on the graphic and morphological structure, and hence on the individual unit level in that scheme.

As comic book pages are complex documents, the analysis, detection and segmentation of their constituent parts is an interesting problem for automated document analysis. Typically a comic book consists of pages divided into panels according to a variety of layout schemes. Panels in turn contain pictorial content as well as several types of text containers such as captions and speech balloons, which usually differ in function. Panels may also contain onomatopoeia and diegetic text. The individual elements just mentioned can also be distributed across several panels or bleed into the void between panels called the ``gutter''. Recent overviews of comics research can be found in~\citet{Cohn:2013cs},~\citet{8270237}, and~\citet{Dunst.2018book}.
Here we focus on one particular element of graphic structure, namely, speech balloons. Most of the text in comics is located in speech balloons and captions, thus locating these elements is a prerequisite for OCR. However, it also pays to classify these as different classes, because they serve different functions: In contrast to captions, which are are normally used for narrative purposes, speech balloons typically contain direct speech or thoughts of characters in the comic. Speech balloons usually consist of a carrier, ``a symbolic device used to hold the text'' (Cohn, 2013b), and a tail connecting the carrier to its root character from which the text emerges. Both tails and carriers come in a variety of shapes, outlines, and degrees of wiggliness. Here we describe a method for the automated detection and segmentation of speech balloons of various forms and shapes, including their tails. 

Our method is based on a fully convolutional network~\cite{Shelhamer2015FullyCN}, specifically on a modified U-Net architecture~\cite{10.1007/978-3-319-24574-4_28} in combination with a pre-trained VGG-16 encoder~\cite{DBLP:journals/corr/SimonyanZ14a}. We are thus using an encoder that was trained for classification of natural images, and an architecture which was originally developed for medical image segmentation. By showing that this combination achieves state-of-the art performance with  illustrations, we also suggest that transfer might generalize to more or less related tasks.
Automated detection of speech balloons and other typical elements of comic books is not only desirable for research purposes such as creating and analyzing larger datasets of annotated comic books, or as a first step in an optical character recognition (OCR) pipeline, but also has potential for applications such as preparing digitized comics for computerized presentation on handheld devices, or developing assistive technologies for people with low vision.

\subsection{Related work} 
\label{ssec:related_work}
Recently, deep convolutional neural networks (DCNNs) using learned feature hierarchies have been shown to perform better in many visual recognition and localization tasks than other algorithms that use carefully crafted engineered features. However, partly due to the lack of big corpora there are only very few publications tackling the task of analyzing comic books using deep neural networks. We will describe some of this recent work on DCNN-based object detection in comics below, but first summarize some of the older work based on engineered features, focusing on speech balloon detection. 

\citet{Arai:2011sf} were the first to attempt speech balloon detection. They limited themselves to simple layouts without overlap or bleeding, i.e., all balloons are fully contained within the panels. They first extracted panels using a connected-components approach. Some morphological operations are then applied to panels, followed by another connected component run, from which candidates are selected using some heuristics such as white pixel ratio or size and height relative to the panel dimensions. Some of these heuristics assume vertical text, so the applicability is limited to Manga.

\citet{6195407} use a similar approach, but start by identifying candidate regions through their (white) color. Next they apply some size and shape criteria, then use connected components to locate text within the balloon, followed by a morphological dilatation operation, and another size threshold. The result is a bounding box of the text. Again, this approach only works with simple balloons.

\citet{DBLP:conf/icdar/RigaudBOKW13}, realizing the great variety and heterogeneity of balloons in the wild, which are often defined by irregular shapes or even illusory contours, developed a decidedly different approach based on active contours~\cite{Kass1988}. Following edge detection by a Sobel operator, and text detection (and removal) as an important initialization step, \citet{DBLP:conf/icdar/RigaudBOKW13} introduce new energy terms for the active contour model based on domain knowledge, such as strong edges, smooth contours, and the relative location of text. This is the first proposal that can delineate the full shape of a speech balloon and handle illusory or subjective contours.

Whereas this method requires the location of text,~\citet{Rigaud:2017zm} propose a method in which speech balloons are located first, and text is only used to give a confidence rating on whether a candidate region is indeed a speech balloon. This method first uses an adaptive binarization threshold, following by a connected components approach to identify (a) speech balloon candidates and (b) text within speech balloons. Some criteria on typical alignment and centering of text within a balloon are used to arrive at a confidence score. One drawback of this method in comparison to~\citet{DBLP:conf/icdar/RigaudBOKW13} is that it only appears to work with closed balloons.

Let us now turn to DCNN approaches. Since we know of only one attempt at speech balloon segementation proper, we extend the scope a little. The general aim is to show that features which are typically learned on photographs of natural scenes can nevertheless be used to describe illustrations such as comics. More specific applications in object detection and segmentation are included.

\citet{Chu:2017:MFF:3078971.3079031} used deep neural networks for face detection in Manga. Rigaud and colleagues, who had previously used several methods with hand crafted features to detect elements of comic book pages (see above), have switched to convolutional neural network features in more recent work~\cite{DBLP:conf/icdar/NguyenRB17,10.1007/978-3-030-05716-9_57}.~\citet{10.1007/978-3-030-05716-9_57} describe a multi-task learning model for comic book image analysis derived from on Mask R-CNN~\cite{DBLP:journals/corr/HeGDG17} that defines the current state-of-the-art in detection and segmentation of several object categories such as panels, characters, and speech balloons.~\citet{Ogawa2018ObjectDF} used convolutional neural network features for object detection on the annotated Manga109 corpus.

\citet{IyyerComics2016} attempted to model a striking sequential aspect of comics, namely the inference required from the reader to bridge the gutter, i.e., to make sense of panel transitions and coherently integrate them into the mental model of the text. The multilevel hierarchical LSTM architecture they used to model sequential dependencies in text and image semantics is beyond the scope of the present paper, but in passing they provided an annotated dataset of comics in the public domain from the ``Golden Age of Comics''. Specifically, they (a) annotated 500 randomly selected pages for training a Faster R-CNN \cite{DBLP:journals/corr/RenHG015}  to segment pages into about 1.2 million panels. Furthermore, they (b) annotated 1500 panels for textboxes for training another Faster R-CNN combined with ImageNet-pretrained VGG feature encoding to segment 2.5 million textboxes, and (c) extract text from these textboxes using Google's Cloud Vision OCR. This is certainly a data set that deserves further exploration. For the current purposes, the main problem is that text elements are annotated using rectangular bounding boxes, and without differentiating between captions and speech balloons. Thus segmentations based on these annotations lack detail and specificity. Generally, however, their results definitely show that pretrained features obtained from training on ImageNet can very well be re-used for comics document image analysis tasks.

~\citet{10.1007/978-3-030-05716-9_61} have similarly shown that deep convolutional neural networks are well-suited to capture graphical aspects of comics books and can successfully be used in categorizing illustrator style, pacing the way for a visual stylometry.~\citet{Laubrock:2019jb} show that eye-tracking data obtained from comics readers as a measure of visual attention and CNN predictions of empirical saliency based on Deep Gaze II~\cite{Kummerer_2017_ICCV} tend to focus on similar regions of comics pages. Regions containing text such as speech balloons or captions are focused most often. For the image part, empirical attention allocation appears to be object-based, with regions containing faces receiving a high proportion of fixations.

\subsection{The present approach} 
\label{ssec:the_present_approach}
Building on these earlier successes, here we use a deep neural network to learn speech balloon localization and segmentation of annotated pages from the Graphic Narrative Corpus~\cite{DBLP:conf/icdar/DunstHL17}. We regarded speech balloon detection as a semantic segmentation problem, i.e., a pixel-wise classification task. By partitioning a scanned image into coherent parts we aimed to classify for each pixel whether it belongs to a speech balloon or not. We chose a fully convolutional encoder-decoder architecture based on U-Net~\cite{10.1007/978-3-319-24574-4_28}, i.e., there are no densely connected layers in the model. The network can be divided into an encoding and a decoding branch, devoted to describing the image using a pre-trained network and semantically-guided re-mapping back to input space, respectively. The encoding branch uses a standard VGG-16 model~\cite{DBLP:journals/corr/SimonyanZ14a} pre-trained on ImageNet~\cite{Deng2009ImageNetAL} that employs several hierarchical convolutional and max-pooling layers to extract features useful in describing the stimulus. It therefore encodes the image into more and more abstract representations guided by semantics. The higher the encoding level, the less fine-grained the spatial resolution and the more condensed the semantic information. We have previously shown that re-using weights obtained from pre-training for an object recognition task transfers well to graphic novels~\cite{10.1007/978-3-030-05716-9_61}. We chose VGG for encoding because of the conceptual simplicity of its architecture that allows for easy upscaling, but note that other architectures that outperform VGG in image classification (such as Xception,~\cite{8099678}) can in principle be plugged in.

The decoding branch of the network performs upsampling and is trained to do the segmentation. The decoder thus semantically projects discriminative features learnt by the encoder onto the pixel space of the input image in order to get a dense classification. Intermediate representations from the encoding branch are copied to and combined with the decoding activations to achieve better localization and learning of features in the upsampling branch.

\section{Methods}
\label{sec:methods}

\subsection{Dataset}
\label{ssec:dataset}
Our training material contains roughly 750 annotated pages of the Graphic Narrative Corpus~\cite{DBLP:conf/icdar/DunstHL17}, representing English-language graphic novels with a variety of styles. Parts of the GNC are annotated by human annotators with respect to several constituent parts such as panels, captions, onomatopoeia, or speech balloons. The material includes some non-standard examples of balloons (e.g., speech balloons outside of the panel or balloons looking like captions) but mainly consists of typical balloons from a large variety of 90 comic books.

Binary mask images were generated from ground truth corpus annotations of speech balloons, which are represented by lists of polygon vertices in an XML format. For the balloon segmentation task, we downscaled the corpus images to a fixed size of $768\times512$ pixels in RGB.

\subsection{Model architecture} 
\label{sub:model_architecture}
\begin{figure*}
  \centering
  \includegraphics[width=.9\textwidth]{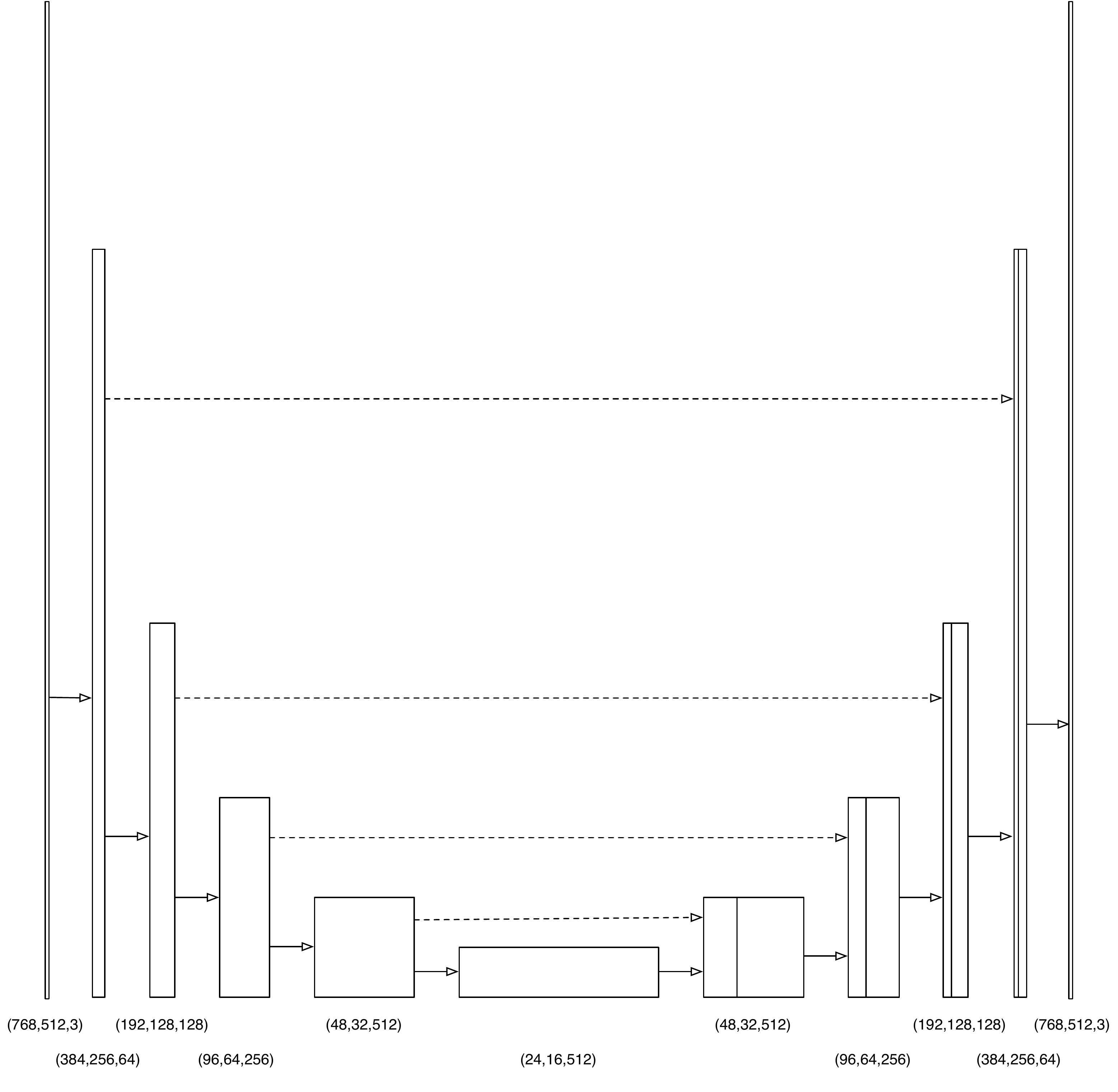}
  \caption{Overview of the model architecture. The encoding branch on the left processes the input image of a comic book page and generates and abstract and increasingly semantic description of the image in terms of VGG16 features, whereas the decoding branch on the right combines upsamples of internal representations with copies from the encoding branch to ultimately arrive at the prediction of a mask image. Box height (not drawn to scale) represents the product of x and y dimensions, and box width (not drawn to scale) represents the number of channels.}
  \label{fig:fig1}
\end{figure*}

We used a fully convolutional approach predicting a pixel-based image segmentation. The network is based on the U-Net architecture and can be divided into an encoding and a decoding branch (see Figure~\ref{fig:fig1}). 

The encoding branch uses the convolutional part of the VGG-16 model. Each image is downsampled subsequently to five encoding representations, each halving the spatial height and width of the previous representation. This is simply achieved by using the output of the five convolutional blocks of the VGG-16 model (Figure~\ref{fig:fig1}, left side).

The decoding branch decodes up to the original resolution. Each of the five upsampling steps (Figure~\ref{fig:fig1}, right side) doubles the width and height of an encoding representation. Upsampling is performed by transposed convolutions. At each upsampling step, skip connections are used, meaning that copies of the encoding representation corresponding to the resulting width and height are additionally included by concatenation. A convolutional operation was performed on the concatenated representation. A sigmoid activation function as the last step of decoding results in a $768\times512$ representation of floating values in (0,1), which can be considered to give the probability of each pixel to belong to a speech balloon. The decoder thus outputs a pixel-based prediction about the location of the speech balloons, which was compared to binary mask images of our annotated speech balloons.

In addition to the standard U-Net architecture, in the upscaling branch, we used L2 kernel-regularization (see results for parameters) and standard batch normalization~\cite{DBLP:journals/corr/IoffeS15} after the last activation function of each upsampling step. We implemented the model using keras~\cite{chollet2015keras}.

\subsection{Loss function} 
\label{sub:loss_function}
We used the sum of the pixel-wise binary cross entropy loss as a sum over each pixel $i$ in the set of $N$ pixels
\begin{equation}
	L_{bce} = -\frac{1}{N}\sum_{i}{(y_i \log{\hat{y_i}} + (1 - y_i) \log{(1 - \hat{y_i})})}
\end{equation}
 and the Dice loss 
\begin{equation}
	L_{Dice} =1 - \frac{2 \sum_{i}{y_i \hat{y_i}}}{\sum_{i}{y_i^2}\sum_{i}{\hat{y_i}^2}}
\end{equation}
of the model prediction and the binary image mask of the balloon annotations as loss function for training.


\subsection{Training and augmentation} 
\label{sub:training_and_augmentation}
We split our dataset into a training and validation set with the ratios 0.85 and 0.15 on the total set, respectively. This split was done randomly with a stratifying condition ensuring to split each of the included comics books with this ratio (some comic books were pooled due to the low amount of annotated images).
First, all image values were linearly normalized to the interval [0,1]. Then, on the training set we applied image augmentation randomly using rotation of the hue channel of the HSV image (range 0 to 0.3), height and widths shifts (range: 0.2 of each dimension size) and flipping the image horizontally and vertically. We trained the model using an Adam optimizer (learning rate of $0.001$, $beta_1=0.9$, $beta_2=0.999$) for a total of 500 epochs, with each epoch going through the full training set. In order to fight overfitting, weight decay (L2 regularization) with a parameter of~$ \lambda=0.001$ was added to the final convolutional layer of each decoding block, which had an effect similar to dropout.


\section{Results} 
\label{sec:results}
First, we report the results of five different splits of training and validation data. Due to the regularization term in addition to random fluctuations, results can vary from epoch to epoch; in order to obtain a more representative value we smoothed the values by computing the median of the metrics on the validation set for the last five epochs. The values for binary cross entropy loss, the Sørensen-Dice coefficient, precision and recall are reported in Table~\ref{tab:table_metrics}.

\begin{table*}[htbp]
    \centering
	\begin{threeparttable}[htbp]
	\caption{Validation metrics on GNC.}
	\label{tab:table_metrics}
\begin{tabular}{ l l l l l  l l l l}
\toprule
& \multicolumn{4}{c}{Training}& \multicolumn{4}{c}{Test}\\
Run & BCE\tnote{1} & Dice\tnote{2} & Recall & Precision & BCE\tnote{1} & Dice\tnote{2} & Recall & Precision \\
\midrule
1 & 0.257 & 0.976 & 0.976 & 0.980 & 0.899 & 0.940 & 0.931 & 0.955 \\
2 & 0.269 & 0.975 & 0.975 & 0.979 & 0.838 & 0.941 & 0.941 & 0.947 \\
3 & 0.271 & 0.975 & 0.975 & 0.979 & 0.711 & 0.949 & 0.943 & 0.960 \\
4 & 0.343 & 0.969 & 0.970 & 0.974 & 0.649 & 0.951 & 0.952 & 0.957 \\
5 & 0.275 & 0.974 & 0.975 & 0.978 & 0.788 & 0.943 & 0.935 & 0.960 \\
\midrule
average & 0.309 & 0.974 & 0.974 & 0.978 & 0.777 & 0.945 & 0.940 & 0.956\\
\bottomrule
\end{tabular}
\begin{tablenotes}
         \item [1] Binary cross entropy loss
		 \item [2] Sørenson-Dice coefficient
         \end{tablenotes}
\end{threeparttable}
\end{table*}


\begin{table*}[ht!]
    \centering
	\begin{threeparttable}[ht!]
	\caption{Speech balloon segmentation performance in percent.}
	\label{tab:table_comparison}
\begin{tabular}{ l l l l}
\toprule
Method & Recall & Precision & $F_1$-Measure\\
\midrule
\citet{Arai:2011sf} & 18.70 & 23.14 & 20.69\\
\citet{6195407} & 14.78 & 32.37 & 20.30\\
\citet{DBLP:conf/icdar/RigaudBOKW13} & 69.81 & 32.83 & 44.66\\
\citet{Rigaud:2017zm} & 62.92 & 62.27 & 63.59\\
\citet{10.1007/978-3-030-05716-9_57}, Mask R-CNN & 75.31 & 92.42 & 82.99\\
\citet{10.1007/978-3-030-05716-9_57}, Comic MTL & 74.94 & 92.77 & 82.91\\
Our method, eBDtheque & 75.19 & 89.05 & 78.42 \\
Our method, eBDtheque cleaned w/o Manga\tnote{1} & 82.94 & 91.30 & 85.84\\
Our method, eBDtheque\tnote{2} & 86.79 & 97.81 & 90.22\\
Our method, GNC\tnote{3} & \textbf{94.04} & \textbf{95.58} & \textbf{94.48}\\
 \bottomrule
\end{tabular}
\begin{tablenotes}
	\item [1] excluding Manga and a few ground truth images in which annotated balloons were captions according to our definition
	\item [2] median
	\item [3] Graphic Narrative Corpus test set
\end{tablenotes}
\end{threeparttable}
\end{table*}

To put our results into perspective, we compare the values with previous work by adapting a table from Nguyen et al. (2019). These results were obtained from predicting on the eBDtheque data set~\cite{eBDtheque2013}, which in its version 2 includes pixel-based ground truth. In order to facilitate comparison, we also computed predictions of our model (trained on the GNC) for the eBDtheque corpus. Table~\ref{tab:table_comparison} lists the values for recall, precision, and $F_1$ score (we used the average over five different runs on different test sets). Note that we did not use the eBDtheque material for training, whereas the other models did. Therefore, in order to better describe our model performance, we additionally present results obtained for our own test set of pages from the GNC that were excluded from training (Table~\ref{tab:table_comparison}, last row).

Our model generally transfers quite well to the new data set. One problem is that eBDtheque annotations do not distinguish between speech balloon and caption, whereas our model has learned to distinguish between them. We thus manually added this distinction to the ground truth. Another problem for our model is the inclusion of a few Manga images with Japanese script in the eBDtheque. We had previously observed that our model currently fails miserably to predict speech balloons in the Manga109 data set~\cite{Matsui2017}.  Whereas we can currently only speculate that this may be due to the some sort of basic alphabetic character recognition and/or vertical orientation of the speech balloons, it gave us an a priori reason to exclude the 6 manga images from evaluation, which boosted the score quite a bit. Finally, we observed that out model did very well on the vast majority of images, but performed poorly on a few images with rather untypical speech balloons (like the very abstracted balloons in TRONDHEIM\_LES\_TROIS\_CHEMINS\_\_00x.jpg). The failure to capture these extreme examples is due to a relative lack of diversity of our training set. To get a more representative measure of the performance, we therefore additionally included the median per measure over images. In summary, our model outperforms previous approaches in recall and $F_1$, and is at least competitive in precision.

\subsection{Qualitative results} 
\label{sub:qualitative_results}

\begin{figure*}[htbp]
  \centering
  \includegraphics[width=.76\textwidth]{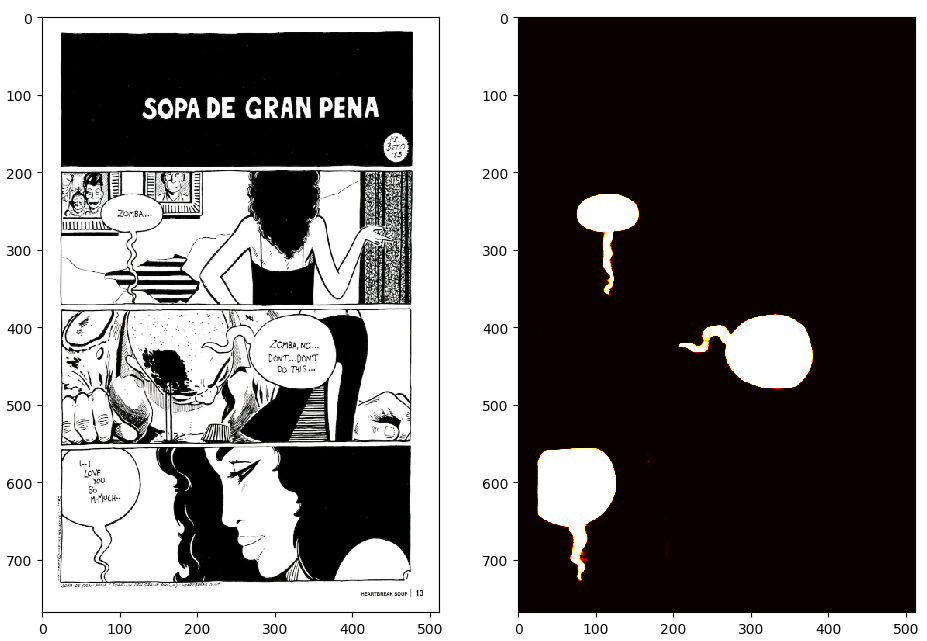}
  \caption{An example from the training set~\citep[p. 13]{Hernandez:2007mi} shows that curved tails are well approximated by the model.}
  \label{fig:fig2}
\end{figure*}

\begin{figure*}[htbp]
  \centering
  \includegraphics[width=.76\textwidth]{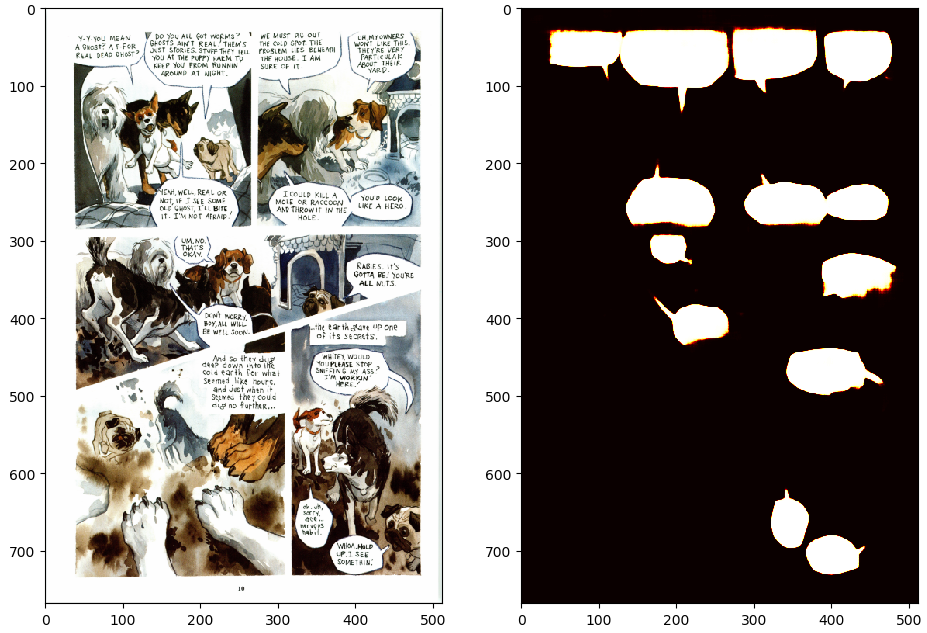}
  \caption{Example from the training set~\citep[p. 10]{dorkin2010beasts} showing that the model can detect and complete illusory contours.}
  \label{fig:fig3}
\end{figure*}

\begin{figure*}[h!]
  \centering
  \includegraphics[width=.76\textwidth]{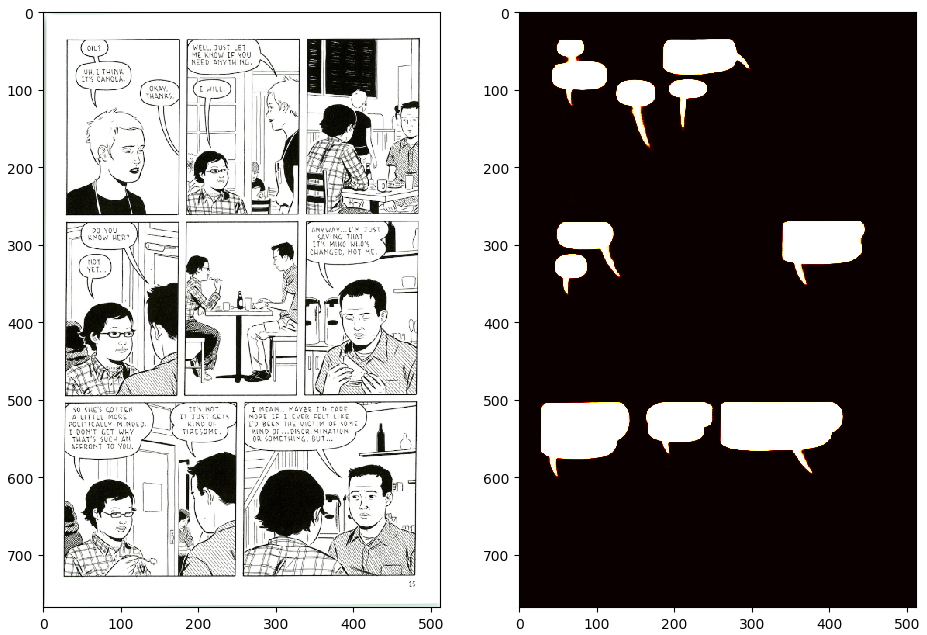}
  \caption{Example from the test set~\citep[p. 15]{Tomine:2007vb} demonstrating decent transfer to unseen material.}
  \label{fig:fig4}
\end{figure*}

\begin{figure*}[h!]
  \centering
  \includegraphics[width=.76\textwidth]{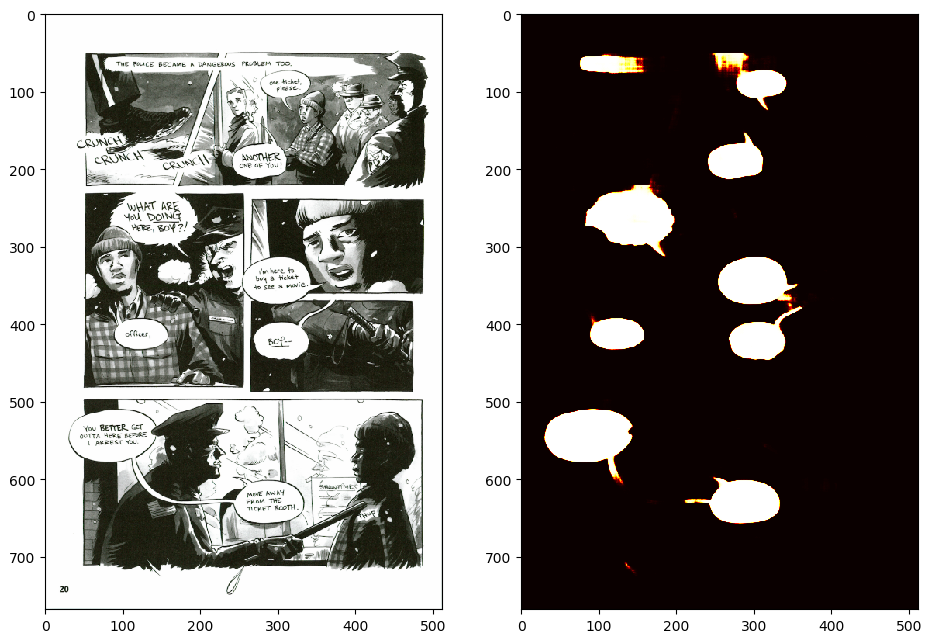}
  \caption{Example from the test set~\cite[p. 20]{Lewis:2015xc} showing that the model sometimes has problems with illusory contours.}
  \label{fig:fig5}
\end{figure*}

Figures~\ref{fig:fig2}-\ref{fig:fig5} show some qualitative results, with colours indicating confidence of speech balloon classification (values according to matplotlib's~\cite{Hunter:2007} colormap hot, with black: 0, white: 1). Figures~\ref{fig:fig2} and~\ref{fig:fig3} are from the training set, whereas the remaining Figures are from the test set. Figure~\ref{fig:fig2} illustrates that the model can (a) well approximate curved tails, and (b) has learned to differentiate between speech balloons and other regions containing text. Figure~\ref{fig:fig3} demonstrates that the model can approximates illusory contours, i.e., boundaries of speech balloons that are not defined by physical lines, but just by imaginary continuoations of the gutter lines in this case. Classification confidence drops near those illusory contours. Figure~\ref{fig:fig4} shows that good performance is also achieved on test set images. Figure~\ref{fig:fig5} (upper part) illustrates that sometimes illusory contours are hard to detect for the model. Furthermore, a false alarm in the lower part of the figure is probably triggered by a shape that looks similar to a tail of a speech balloon.




\section{Discussion} 
\label{sec:discussion}
We obtained state-of-the-art segmentation results for speech balloons in comics by training a fully-convolutional network model based on the U-Net architecture combined with VGG-16 based encoding. The model was able to fit and predict subtle aspects of speech balloons like the shape and direction of the tail, the shape of the outline, or even the approximate end of speech bubbles that were partly defined by illusory contours (Kanizsa, 1976).

Whereas the overall performance is impressive, inspection of misclassification cases is instructive. Most of these appear to be solveable by some heuristic post-processing using methods similar to those described in~\ref{ssec:related_work}. Additionally, a relatively high confidence threshold would help reduce the number of false alarms. For example, sometimes the model is weakly confident about having detected a balloon in a bright area that has a boundary akin to a circular arc. However, the confidence for these regions is usually lower than for real balloons, and the detected area is also significantly smaller. Thus, some thresholding on confidence as well as size might be helpful for production use, even though less desirable in terms of using a unified approach. Another class of misclassification is misses (omissions); inspection of these suggest that they were often ballons with very large fonts, and sometimes jagged boundaries. Since very large typeface is otherwise mostly used for onomatopeia, we think it is possible that the model has learned that large fonts are usually associated with something else than speech balloons.

One limitation of the current model is related to the training material. In testing transfer to new data sets we observed that whereas predictions for the eBDtheque data set were qualitatively good, the model in its current form failed quite miserably on the Manga109 data set~\cite{Matsui2017,Ogawa2018ObjectDF}. We speculate that the model currently encodes some culture-specific features such as letter identities of the latin alphabet, the horizontal orientation of text lines, or the horizontally elongated shape of most speech balloons in our training material. An updated version that includes more Manga training material is under development.

Our original aim for successful detection was to integrate the semi-automated speech balloon detection in our in-house annotation tool (Dunst et al., 2017). Even though prediction times are pretty fast, they will introduce a noticeable delay in page loading, thus a batch mode is planned. Regions that are predicted to be speech balloons will be represented by their polygonal outline, which can then be accepted or corrected by the annotator. Alternatively, means of optimizing the run time per image will be explored.

Another obvious for future research is to extend the model to allow segmentation of different classes, such as captions, onomatopoeia, and even characters. The general success of segmentation model in multiclass labeling as well as the specific results reported by Nguyen et al. (2019) in their multitask model suggest that this will be quite feasible.
A more general use of this kind of algorithm (using different training data) would be any segmentation task using real-world features. For example, we expect that the present approach may generalize to other segmentation tasks such as segmenting historical manuscripts into text and image, scientific articles into text, figures and tables, segmenting newspaper articles into paragraphs and images, or segmenting medical scans according to presence or absence of a disease. Of course, human-assisted verification is needed, but given the fact there are now several computer vision domains where the performance of CNN-based models is at least close to human performance, we expect to be able to solve several tedious annotation tasks, freeing human resources for more interesting endeavours.

\section*{Acknowledgment}
This work was funded by BMBF grant 01UG01507B to JL. The authors would like to thank Christophe Rigaud and the eBDtheque team for making available their database.

\nocite{DBLP:conf/mmm/2019-2}
\pdfbookmark[1]{References}{References}
\bibliographystyle{plainnat}  


%
%
%
%

\end{document}